\documentclass[runningheads]{llncs}
\usepackage[T1]{fontenc}
\usepackage{graphicx}
\begin{document}
%
\title{XCB: an effective contextual biasing approach to bias cross-lingual phrases in speech recognition}
%
%
\author{Xucheng Wan, Naijun Zheng, Kai Liu, Huan Zhou}
\authorrunning{X. Wan et al.}
%
\institute{IT Innovation and Research Center, Huawei Technologies Co., Ltd. \\
\email{\{wanxucheng,zhengnaijun,liukai89,zhou.huan\}@huawei.com}}
\maketitle
\begin{abstract}

Contextualized ASR models have been demonstrated to effectively improve the recognition accuracy of uncommon phrases when a predefined phrase list is available. However, these models often struggle with bilingual settings, which are prevalent in code-switching speech recognition. 
In this study, we make the initial attempt to address this challenge by introducing a Cross-lingual Contextual Biasing(XCB) module. Specifically, we augment a pre-trained ASR model for the dominant language by integrating an auxiliary language biasing module and a supplementary language-specific loss, aimed at enhancing the recognition of phrases in the secondary language. 
Experimental results conducted on our in-house code-switching dataset have validated the efficacy of our approach, demonstrating significant improvements in the recognition of biasing phrases in the secondary language, even without any additional inference overhead. Additionally, our proposed system exhibits both efficiency and generalization when is applied by the unseen ASRU-2019 test set. 

\end{abstract}
\section{Introduction}

For the past decade, End-to-end (E2E) ASR models have demonstrated remarkable progresses on speech recognition. Driven by large scale dataset, typical E2E Models like Transformer\cite{attention,transformerASR}, Transducer\cite{transducer,convTT} and Conformer\cite{conformerASR} have been reported to yield the best results to-date on variant speech recognition tasks. Nevertheless, these E2E ASR systems might encounter recognition errors with long-tailed rare words (such as jargon or unusual named-entities in unique target domain), presenting a challenge for real-world ASR implementations.


To address this issue, one popular way is to enhance an E2E ASR system by integrating contextual information (extracted from a predefined hotword list of rare words), known as contextualized ASR\cite{shallow1,shallow2,deepcontext1,deepcontext2,dataaug}. For example, Paraformer\cite{paraformer}, as a non-autoregressive (NAR) model, has recently attracted increasing attention due to its high accuracy and efficient NAR inference. To enable it with hotword customization ability, SeACo-Paraformer\cite{seaco} was proposed with improved ASR accuracy and hotwords recall rate. Nevertheless, in practical scenarios with code-switching speech, these contextualized ASR systems, typically trained on one specific language data, often struggle with those hotwords of secondary language. 

On the other hand, in code-switching ASR domain, various efforts are made with objective of learning language-specific representations and discriminate boundaries between different languages. Representative works include employing language expert modules (mixture of experts, MOE) with multi-encoders \cite{bilingual,multi-encoder,bi-encoder}, language-aware encoder \cite{lae1,lae}, and separate language-specific representations with adapters \cite{ba-moe}. While using code-switching ASRs to recognize the cross-lingual phrases is feasible, it typically requires a large-scale training data, which is impractical in most cases.  

In all, to our best knowledge, the challenge for contextualized ASR to bias cross-lingual phrases has not be addressed in prior arts. Therefore, in this study, we make the first attempt to address this challenge by proposing a Cross-lingual Contextual Biasing(XCB) enhanced ASR system. Inspired by the concept of MOE, we developed a supplementary lightweight XCB module, integrated with a pre-trained contextualized ASR model. By leveraging the XCB module and an additional training loss item, we aim to improve the learning of representations and boundaries for the secondary language (L$_{2nd}$), within utterance-level acoustic embeddings. This approach enables the XCB-enhanced system to achieve a performance boost on biasing phrases in L$_{2nd}$, with minimal training data and negligible computational overhead during inference. 

In the rest of the paper, we first review the contextualized ASR backbone (SeACo-Paraformer) in Section 2, and describe our XCB-enhanced backbone in Section 3. Experimental results are presented in Section 4 and Section 5 concludes the paper.

\section{Preliminary}
\subsection{Paraformer}
As a fast and accuracy parallel transformer, Paraformer\cite{paraformer} is a powerful NAR(non-autoregressive) ASR model trained with a large amount (more than 20k hours) of data. To speed up inference, two key technologies are designed. One is a continuous integrate-and-fire-based (CIF) based predictor to accurately predict the number of output tokens, the other is a glancing language model (GLM) based sampler to enhance the NAR decoder with the ability to model token inter-dependence. Benefiting from the technologies, on many public Mandarin benchmarks, Paraformer delivers remarkable performance on par with state-of-the-art AR systems, with more than a 10-fold speedup.

\subsection{SeACo-Paraformer}
SeACo-Paraformer\cite{seaco} integrates the ability of hotword customization to the Paraformer backbone. By appending a semantic-augmentation contextual module aside the encoder, SeACo extracts hotword embeddings from the hotword list and sends them into the bias decoder to obtain biased acoustic embeddings and biased semantic embeddings, and finally yielding contextual biased probabilities. 
Due to its high accuracy and efficiency, in this study, we adopt the SeACo-Paraformer as our backbone for contextualized ASR system.

\section{Proposed Methods}
To bias cross-lingual phrases in context of SeACo-Parafromer (pre-trained on large-scale first language(L$_{1st}$), we introduce an additional XCB module and a language specific loss component, with motivation to enhance the acoustic embeddings associated with L$_{2nd}$.  
\subsection{XCB Module}
\begin{figure}
\includegraphics[width=\textwidth]{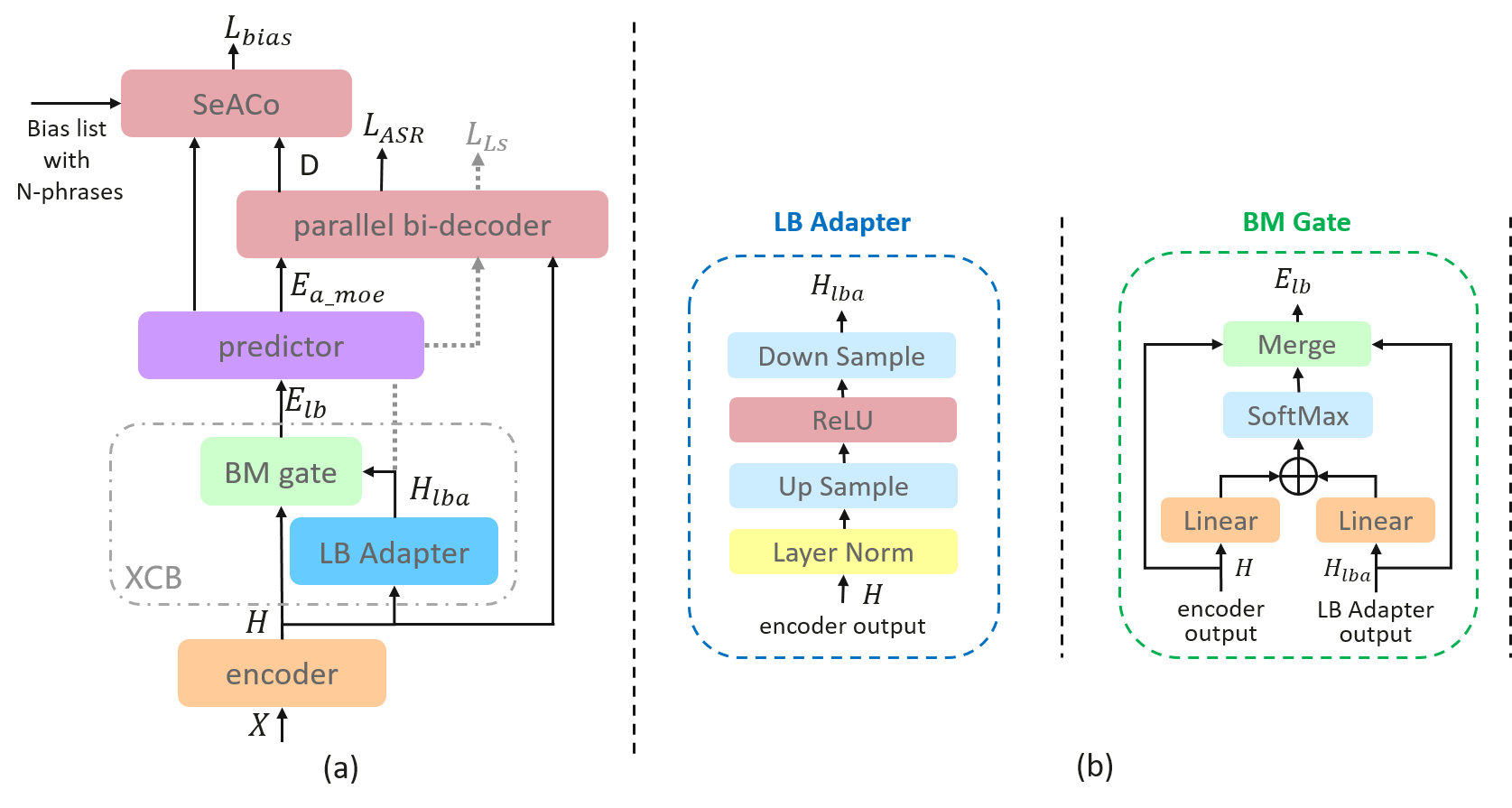}
\caption{Illustration of the proposed XCB-enhancement on the SeACo-Paraformer: (a) the overall architecture; (b) detailed structure of LB Adapter and BM Gate components.} \label{fig1}
\end{figure}

The proposed XCB module is sandwiched between Paraformer encoder and predictor, as illustrated in Fig.\ref{fig1}(a). It comprises two core components: Language Biasing Adapter(LB Adapter) and Biasing Merging Gate(BM Gate), which are elaborated in Fig.~\ref{fig1}(b).

\noindent\textbf{LB Adapter} 
LB adapter is piled up with up- and down-sampling layers, interleaved with layer normalization and ReLU activation. It transforms $H$ (the hidden representations of acoustic features $X$) into a language-biased hidden representation $H_{lba}$.
The LB Adapter serves to distinguish the frames that associated with L$_{2nd}$ and enhance the corresponding representations in feature space of $H_{lba}$. To uphold our backbone's consistent performance on dominant language (L$_{1st}$), we refrain from designing a dual adapter for L$_{1st}$. This approach distinguishes our method different from conventional MOE-based architecture used in code-switching ASR.

\noindent\textbf{BM Gate}
The BM gate takes in hidden representation $H$ and language biased representation $H_{lba}$ to generate language-biased acoustic embedding $E_{lb}$. The dual path inputs are fed into linear projection separately to obtain the language-specific weights, then scaled by these weights, respectively. Lastly, both scaled results and raw input representations (with residual connections) are merged to yield embedding $E_{lb}$. 
The whole process can be formulated as following:
\begin{equation}
    E_{lb}=BMGate(H,\:LBAdapter(H))
\end{equation}

\subsection{Language Specific Loss}
To encourage the learning of the language-biased acoustic embeddings $E_{lb}$, an additional loss component, $L_{2nd}$, is introduced to be combined with the original end-to-end system training loss (joint ASR and biasing loss). 
For this purpose, a L$_{2nd}$ GT label set is firstly constructed. This involves masking the L$_{1st}$ context with <unk> tokens and retaining only the L$_{2nd}$ context in each GT label. Then, in parallel to normal ASR decoding process, another decoding branch is built (the grey dotted line presented in Fig.\ref{fig1} (a)) to explicitly predict the L$_{2nd}$ tokens, as illustrated in Fig.\ref{fig1}(a). The language-biased representation $H_{lba}$ is fed into the predictor to produce 
CIF-triggered L$_{2nd}$ tokens, then these tokens are directly fed into the decoder to produce posterior probabilities $P_{L_{2nd}}$, which is used to calculate loss $L_{CE}^{L_{2nd}}$. The cross-entropy (CE) loss calculated between $P_{L_{2nd}}$ and the L$_{2nd}$ GT labels is referred to as $L_{CE}^{2nd}$. Overall, the total loss function is structured as a linear combination of three loss components, formulated as:
\begin{equation}
L_{total} = L_{ASR} + L_{bias} + \alpha L_{CE}^{2nd}
\end{equation}
\noindent where $\alpha$ is a hyper-parameter determining the contribution of proposed language-specific loss.

\section{Experiments}

\subsection{Data Preparation}
Our proposed ASR model with XCB-module is trained and evaluated on an internal code-switching (Mandarin-English) data. This in-house dataset has 13 independently recorded industrial audio data, containing 14k utterances with duration of 20 hours. All these recordings are code-switching utterances, collected with diverse speech topics and recording environment.

The in-house data is randomly split, with 11 recordings for training and 2 for testing. 
Considering that most early works on code-switching ASR use the ASRU 2019 Mandarin-English code-switching challenge dataset, we also perform evaluation experiment on the ASRU dataset. Note that since it is not publicly available, we can only access its test-set (about 16K utterances). Due to limited data resource, we can not provide comparisons with those prior arts, but only to gauge the generalizability and robustness of our proposed model. That is, to apply the XCB model trained on our in-house dataset to the unseen ASRU test-set without any finetuning.

To prepare the hotword list, for each code-swtiching utterance, $N$ bi-lingual phrases are selected. Besides one target L$_{2nd}$ phrase, the remaining $N-1$ phrases are selected from the whole test hotword list, composed of all extracted named-entities (either in L$_{1st}$ or L$_{2nd}$).  of size around 60 for each testing utterance. The English word and Mandarin key entities of the utterance is selected as target hotwords, and they are mixed with interference items collected from contextual contents of the same audio recording to form the hotword list.


\subsection{Experimental Setup}
The official published model \footnote{\url{https://github.com/modelscope/FunASR}} of SeACo-Paraformer is used as our ASR baseline system, which was pre-trained using up to 50k hours Mandarin speech data. Initialized by the baseline, our proposed XCB-enhanced ASR model is trained using the in-house training dataset for 10 epochs, with learning rate equalling 0.0002 and batch size of 30. The weight $\alpha$ is set as 0.3 and no averaging action is applied to model checkpoints.

Apart from the conventional mixed error rate (MER), three more evaluation metrics are adopted to measure the performance on biasing phrases. They are: biasd character error rate(BCER) for Mandarin, biasd word error rate(BWER) for English and biasd mixed error rate(BMER) representing overall biasing performance, defined as:
\begin{equation}
BMER=\frac{n_{bc}*BCER+n_{bw}*BWER}{n_{bc}+n_{bw}}
\end{equation}
\noindent where $n_{bc}$ denotes the number of biasd Mandarin characters and $n_{bw}$ is the number of biasd English words.

\subsection{Experimental Results}
The experimental results are presented in Table\ref{tab1}. Besides the SeACo baseline, we also directly finetuned SeACo using the in-house training dataset, and add the supervised-finetuned version (termed SeACo:sft) as another baseline for performance comparison. 

From the experiment results, we obtain a few valuable insights: 
1) as expected, our proposed XCB system significantly outperforms both baselines in terms of BWER. On the in-house test dataset, our system shows an impressive 17.2\% relative reduction comparing to the vanilla backbone, and 8\% over the fine-tuned backbone; 
2) comparing SeACo and XCB, slight performance differences are observed regarding the general MER and BCER, these confirm that our proposal does not compromise the existing recognition capability on the dominant language; 
3) our proposed method improves both precision and recall of biased English phrases, with significant absolute gains over 9\%; 
4) the results of extending our approach to the unseen ASRU test-set without further fine-tuning on ASRU data, also show considerably performance boost, revealing the efficiency and generalization of our proposal. 
Lastly, on the ASRU test-set, the performance degradation is noted with comparison between the baseline and its fine-tuned version. It may caused by over-finetuning on the relatively small data. In contrast, without any fine-tuning, our proposed XCB systems achieves the best performance in terms of BWER and BMER.

\begin{table}
\caption{Experimental results on both in-house and ASRU-2019 test-sets.}\label{tab1}
\begin{tabular}{c|c|c|c|cc|cc}
\hline
\hline
Test-set    &   Model             &  MER  & BMER    & \multicolumn{2}{c|}{BCER}  &  \multicolumn{2}{c}{BWER}  \\
\hline
 &&\multicolumn{2}{c|}{Error Rate(\%)} & \multicolumn{4}{c}{ Error Rate(\%) \:\: Precision/Recall(\%) } \\
\hline

&   SeACo    &  13.55    &  9.53    & \: 5.57 \:\: & 95.31/94.58 \:    & \: 53.88 \:\: & 47.79/46.55  \\
 
in-house    &   \:SeACo:sft\:      &  12.58    &  \textbf{8.51}    & \: \textbf{4.94} \:\: & 95.86/95.35 \:  & \: 48.49 \:\: & 54.04/51.94  \\

&   XCB (ours)           &  12.61    & \textbf{8.52}     & \: 5.31 \:\: & 95.51/94.93 \: & \: \textbf{44.61} \:\: & 58.60/55.82 \\

\hline
&   SeACo       &  \textbf{4.63}    &  4.81   & \: \textbf{1.95} \:\: & 98.21/98.07 \: & \: 10.36 \:\:  & 89.92/89.64  \\
 
ASRU   &   \:SeACo:sft\:       &  5.23    &  5.83   & \: 2.39 \:\: & 97.77/97.64 \: & \: 12.51 \:\:  & 87.77/87.49 \\

&   XCB (ours)     &  4.85    &  \textbf{4.71}   & \: 2.02 \:\: & 98.12/98.00 \: & \: \textbf{9.94} \:\:   & 90.34/90.06 \\

\hline
\hline
\end{tabular} 
\end{table}


\subsection{Active v.s. Inactive XCB}
As proved above, our proposed XCB system, by adding auxiliary XCB module on the ASR backbone, outperforms its baseline in both BMER and BWER. Herein, we conduct one more experiment by keeping the auxiliary XCB module inactive during the inference, with expectation to compare our system with the baseline at the condition of same computational complexity. The results are presented in Table\ref{tab2}, where XCB:nBM refers to the inference system that inactivates the XCB module and directly feeds the hidden representation $H$ into the predictor. Interestingly, even bypassed the XCB module, the XCB:nBM system shows even better performance than the XCB, on both test datasets. It sounds appealing considering the XCB:nBM incurs no additional computational overhead. We speculate that our XCB training might encourage discrimination of L$_{2nd}$ in the lower features produced by the encoder. Further in-depth investigation is needed in our future work. 



\begin{table}
\caption{Results of XCB activation study on in-house and ASRU-2019 test-set.}\label{tab2}
\begin{tabular}{c|c|c|c|cc|cc}
\hline
\hline
Test-set    &  Methods             &  MER  & BMER    & \multicolumn{2}{c|}{BCER}  &  \multicolumn{2}{c}{BWER}  \\
\hline
 &&\multicolumn{2}{c|}{Error Rate(\%)} & \multicolumn{4}{c}{ Error Rate(\%) \:\: Precision/Recall(\%) } \\
\hline

\:in-house\:    &   XCB             &  12.61    & 8.52     & \: \textbf{5.31} \:\: & 95.51/94.93 \: & \: 44.61 \:\: & 58.60/55.82 \\

&   \:XCB:nBM\:             &  \textbf{12.42}    & \textbf{8.46}    & \: 5.32 \:\: & 95.51/94.85 \: & \:  \textbf{43.75}  \:\: &  58.84/56.68 \\

\hline
\:ASRU\:   &   XCB    &  4.85    &  \textbf{4.71}   & \: 2.02 \:\: & 98.12/98.00 \: & \: \textbf{9.94} \:\:   & 90.34/90.06 \\

&   \:XCB:nBM\:     &  \textbf{4.65}    &  4.72   & \: \textbf{2.00} \:\: & 98.14/98.04 \: & \: 10.01  \:\: &  90.27/89.99 \\
\hline
\hline
\end{tabular}
\end{table}

\section{Conclusion}

To enhance performance of a contextualized ASR, particularly in recognizing phrases in the secondary language within code-switching utterance, we proposed to extend the ASR with a lightweight auxiliary biasing module along with a loss component. Our proposed approach presents several advantages: it offers significant improvement on recognizing bias phrases of the secondary language; delivers consistent performance on recognition of dominant language; provides generalizability cross different dataset. It allows the ASR system to more effectively handle multilingual inputs and code-switching scenarios. Looking ahead, our next steps involve investigating why ASR performs better with an inactive biasing module compared to an active one. Additionally, we plan to expand our method to incorporate other end-to-end ASR backbones.

%
%
%
%

\end{document}